\documentclass[runningheads]{llncs}
\usepackage{graphicx}
\usepackage{tikz}
\usepackage{comment}
\usepackage{amsmath,amssymb} 
\usepackage{color}

\usepackage[accsupp]{axessibility}  

\newcommand{\paperTitle}{ATEAM: Knowledge Integration from Federated Datasets for Vehicle Feature Extraction using Annotation Team of Experts}
\newcommand{\paperKeywords}{Federated datasets, knowledge integration, Team-of-Experts, VMMR, Re-id}

\newcommand{\shortTitle}{ATEAM: Integrated Vehicle Feature Extraction}

\usepackage{endnotes,microtype,xspace,graphicx,fancyvrb,multirow}

\usepackage{caption}
\usepackage{subcaption}
\usepackage{enumitem}

\usepackage{multirow}
\usepackage[normalem]{ulem}
\useunder{\uline}{\ul}{}
\newcommand{\sys}{\textsc{ATEAM}\xspace}
\newcommand{\toe}{team-of-experts\xspace}

\newcommand{\squishitemize}{
\begin{list}{$\bullet$}
	{ \setlength{\itemsep}{0pt}
		\setlength{\parsep}{3pt}
		\setlength{\topsep}{3pt}
		\setlength{\partopsep}{0pt}
		\setlength{\leftmargin}{1.95em}
		\setlength{\labelwidth}{1.5em}
		\setlength{\labelsep}{0.5em} } }

\newcounter{Lcount}
\newcommand{\squishlist}{
	\begin{list}{\arabic{Lcount}. }
		{ \usecounter{Lcount}
			\setlength{\itemsep}{0pt}
			\setlength{\parsep}{3pt}
			\setlength{\topsep}{3pt}
			\setlength{\partopsep}{0pt}
			\setlength{\leftmargin}{2em}
			\setlength{\labelwidth}{1.5em}
			\setlength{\labelsep}{0.5em} } }

\newcommand{\squishalph}{
	\begin{list}{\alph{Lcount}. }
		{ \usecounter{Lcount}
			\setlength{\itemsep}{0pt}
			\setlength{\parsep}{3pt}
			\setlength{\topsep}{3pt}
			\setlength{\partopsep}{0pt}
			\setlength{\leftmargin}{2em}
			\setlength{\labelwidth}{1.5em}
			\setlength{\labelsep}{0.5em} } }

\newcommand{\squishend}{\end{list}}

\begin{document}
\titlerunning{\shortTitle}
\title{\paperTitle}
%
%
\author{Abhijit Suprem\inst{1} \and
Purva Singh\inst{1} \and
Suma Cherkadi\inst{1} \and
Sanjyot Vaidya\inst{1} \and
Joao Eduardo Ferreira\inst{2} \and
Calton Pu\inst{1}
}
\authorrunning{A. Suprem et al.}
%
\institute{Georgia Institute of Technology, Atlanta, USA \\
\email{asuprem@gatech.edu}\and
University of Sao Paulo, Sao Paulo, Brazil
}

\maketitle              

\begin{abstract}
    The vehicle recognition area, including vehicle make-model recognition (VMMR), 
    re-id, tracking, and parts-detection, has made significant progress in 
    recent years, driven by several large-scale datasets for each task.
    These datasets are often non-overlapping, with different label schemas for each task: VMMR focuses on make and model, while re-id focuses on vehicle ID. 
    It is promising to combine these datasets to take advantage of knowledge across datasets as well as increased training data; however, dataset integration is challenging due to the domain gap problem.
    This paper proposes ATEAM, an annotation team-of-experts to perform cross-dataset labeling and integration of disjoint annotation schemas. ATEAM uses diverse experts, each trained on datasets that contain an annotation schema, to transfer knowledge to datasets without that annotation.
    Using ATEAM, we integrated several common vehicle recognition datasets into a Knowledge Integrated Dataset (KID).
    We evaluate ATEAM and KID for vehicle recognition problems and show that our integrated 
    dataset can help off-the-shelf models achieve excellent accuracy on 
    VMMR and vehicle re-id with no changes to model architectures.
    We achieve mAP of 0.83 on VeRi, and accuracy of 0.97 on CompCars.
    We have released both the dataset and the ATEAM framework for public use.
\end{abstract}

\section{Introduction}
\label{sec:intro}

The vehicle recognition area, including vehicle make-model recognition (VMMR), 
re-id, tracking, and parts-detection, has made significant progress in 
recent years, driven by several large-scale datasets for each task~\cite{reid1,reid2,reid3,vmmr1,vmmr2,vmmr3,vmmr4}. 
Each task starts from vehicle feature extraction and specializes on 
desired annotations: VMMR usually uses make and model annotations, 
whereas re-id focuses on vehicle color and type annotations~\cite{datasetsurvey}. 
As such, datasets for each task have self-selected to contain 
these desired (and necessary) feature annotations. 
For example, VMMR datasets, such as CompCars~\cite{compcars} and Cars196~\cite{cars196}, 
usually contain make and model annotations, while re-id datasets 
like VeRi~\cite{veri} and VRIC~\cite{vric} usually contain vehicle ID annotations.

Since vehicle recognition algorithms require feature extraction, 
images from several vehicle dataset can be useful in model 
training~\cite{cross2}. 
In principle, this applies to any annotations and labels in the 
datasets since the images are the important feature source.  
Thus, integrating diverse vehicle datasets to create a training 
dataset for ve-hicle recognition problems becomes an attractive 
option towards increasing training data. 
In turn, this can increase accuracy and improve robustness due 
to diverse image sources and resolutions~\cite{mtml,vric}.

The integration of heterogeneous data sets is a significant 
challenge due to the problem of the (integrated) global 
schema being a strict superset of sub-schemas in component data sets. 
This is the case of datasets specialized on distinct vehicle 
recognition problems. 
For example, a re-id dataset such as VeRi was not designed for 
make-model recognition, and consequently it would not be expected 
to include annotations of make, model, or color, which are unnecessary 
for its original purpose of re-id. 
Manually filling in the missing feature annotations by human 
annotators is expensive, both in cost and time, a problem 
that would only grow as newer, larger datasets are released.

\subsubsection{Knowledge Integration for Vehicle Recognition.}
\begin{figure}
    \centering
    \begin{subfigure}[b]{0.49\textwidth}
        \centering
        \includegraphics[width=\textwidth]{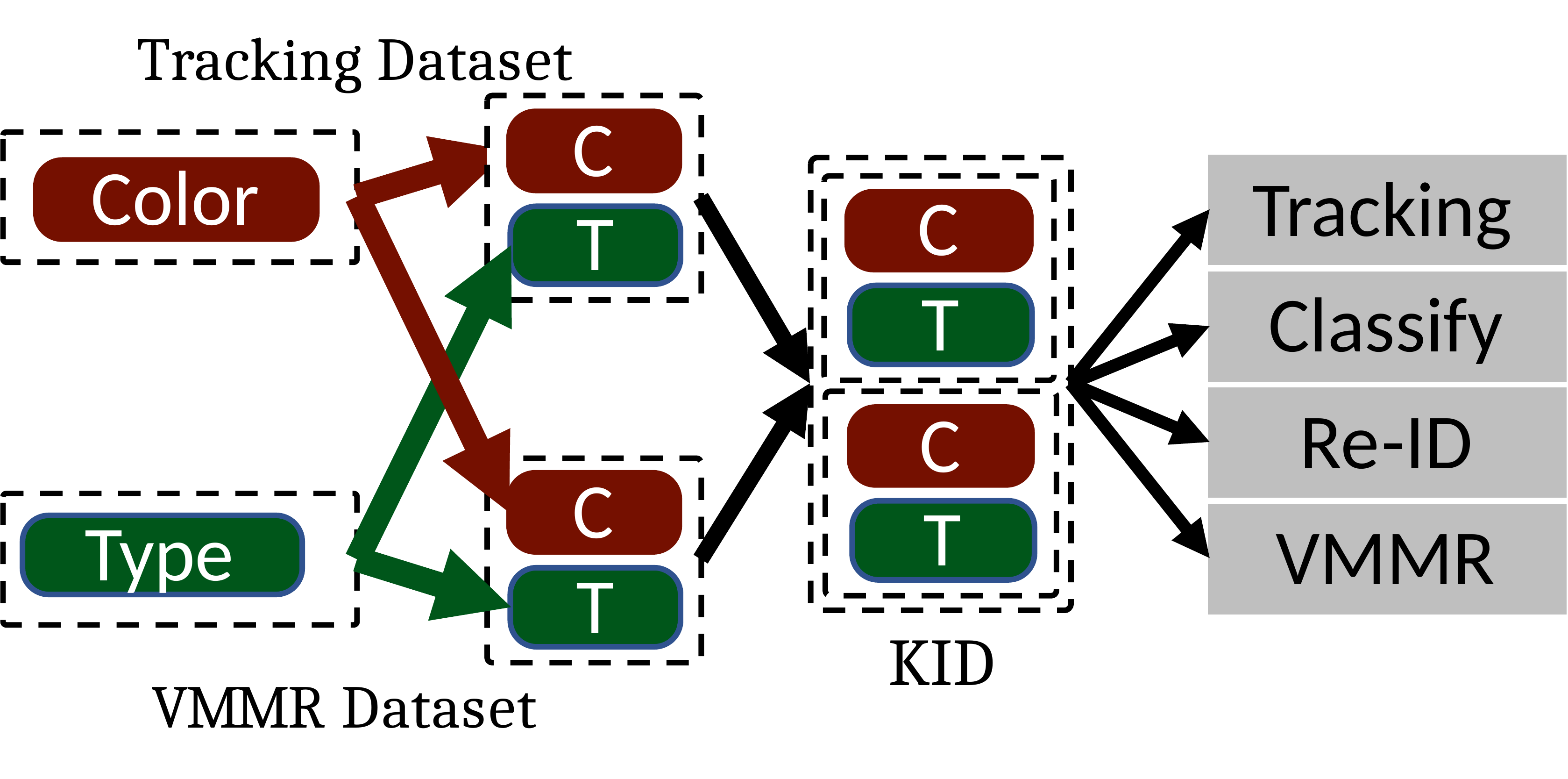}
        \caption{Here, color and type annotations transfer knowledge to each dataset to generate the knowledge-integrated dataset.}
        \label{fig:integratedoverview}
    \end{subfigure}
    \hfill
    \begin{subfigure}[b]{0.49\textwidth}
        \centering
        \includegraphics[width=\textwidth]{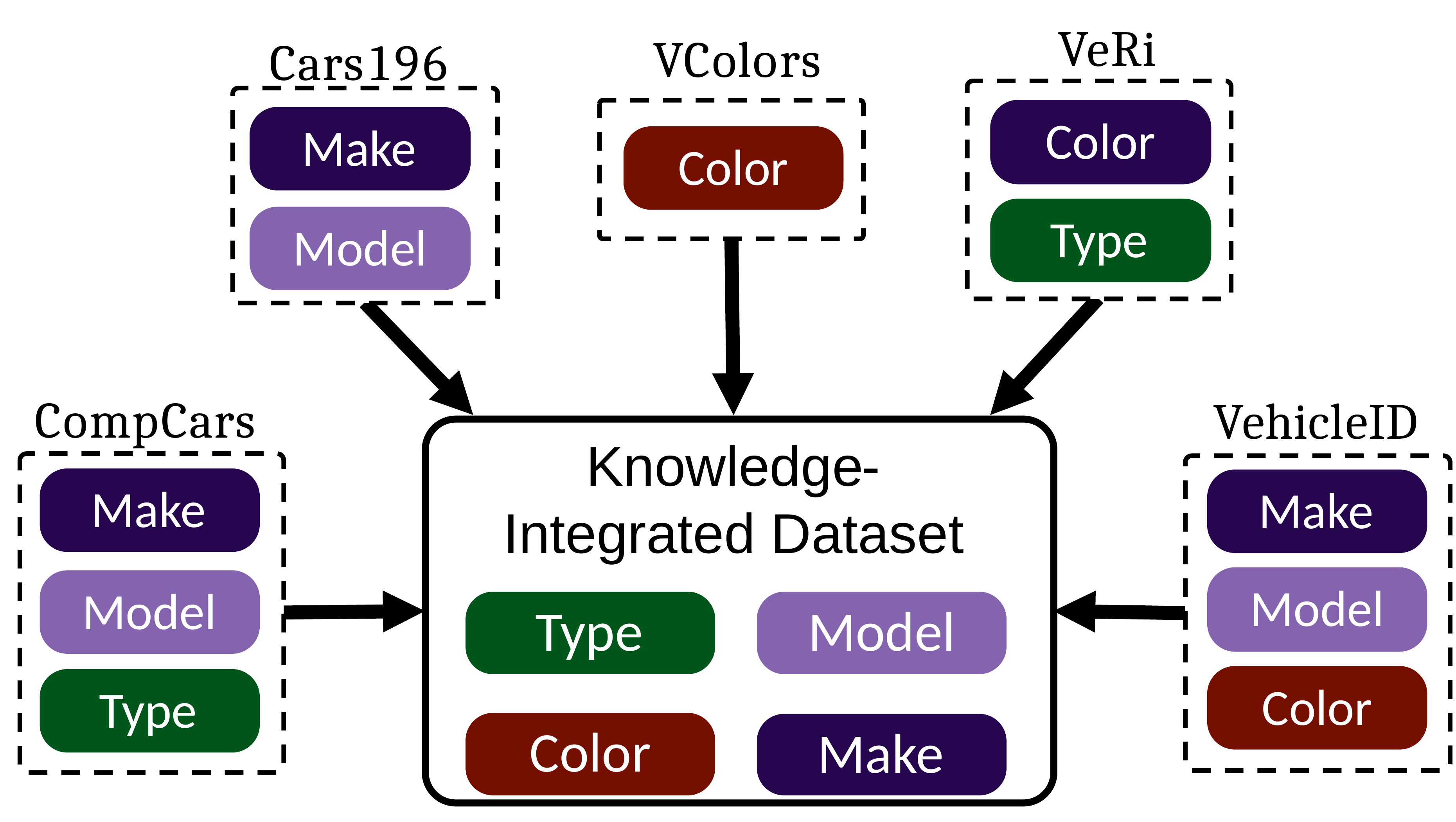}
        \caption{By transferring knowledge between diverse datasets, \sys generates a KID.}
        \label{fig:kid}
    \end{subfigure}
       \caption{Existing vehicle datasets for re-id, classification, or VMMR contain a subset of desired annotations. 
       By propagating knowledge between datasets, we can generate an integrated training dataset with images from all component datasets.}
       \label{fig:overview}
\end{figure}

In the domain of vehicle recognition, fortunately, partial 
knowledge of vehicles from heterogeneous component datasets can 
be integrated into a better understanding of full vehicles. 
A practical solution is to transfer annotation knowledge to 
create knowledge integrated datasets. 
Thus, given federated datasets where a subset has a desired 
annotation, this subset can be used to label datasets without 
the annotation. 
Effectively, the integration process transfers the partial knowledge 
from each component dataset with their own annotations into a 
Knowledge-Integrated Dataset (KID). 
Due to the definition of vehicle features, KID is described by a 
global schema (see Figure~\ref{fig:integratedoverview}) allowing any missing feature annotations 
in component datasets to be inferred by all vehicle recognition tasks. 
%
%
Given two datasets with different label spaces of \textit{color} and 
\textit{type} annotated for different tasks, we train a color model 
to infer color and a type model to infer type labels. 
Once both datasets have color and type labels, we can combine them 
into a KID that can fulfill training requirements of multiple tasks. 
Applying this process repeatedly across several annotations, 
several datasets can be integrated (as in Figure~\ref{fig:kid}) to 
achieve the improvements on vehicle recognition, as we will show in this paper.

\subsubsection{Our Approach.}
In this paper, we propose \sys: a \toe for annotation and building a 
Knowledge-Integrated Dataset for vehicle recognition and show the 
benefits of knowledge integration from component datasets.
In a \toe~\cite{toe1,ebka,ref}, members are trained on different data distributions to 
capture cross-dataset knowledge. 
Our task is to integrate $D$ datasets, each with some subset of the global 
schema with $k$ feature annotations such as make, model, color, and type 
labels, attributes of the global schema. 
The \toe approach transfers knowledge of annotation $k$ from component 
datasets into KID. 
In turn, these improve all vehicle recognition tasks, such as VMMR, re-id, 
or tracking with all $k$ features.

\sys uses $k$ annotation teams, one for each of the $k$ feature annotations 
in $D$. 
The size of the $k$-th team is proportional to the number of datasets 
in $D$ with annotation $k$. 
The $k$-th team labels datasets that do not have the $k$-th annotation. 
We describe the dataflow for color annotation in Figure~\ref{fig:integratedoverview}.

Since each of \sys{'s} teams are trained on one subsets of $D$ and 
used to label a disjoint subset of $D$ without annotations, 
we must address domain gap~\cite{cross1,cross2,cross3,cross4,cross5}. 
Since distributions of datasets are different, some knowledge may 
be lost and may not map exactly during labeling~\cite{odin}. 
This domain gap problem causes deterioration in labeling accuracy 
when samples are significantly different from training data~\cite{litmus}. 
We employ three methods to mitigate the domain gap problem and improve labeling accuracy: 
\textbf{bootstrapping}, \textbf{member confidence}, and \textbf{member agreement}.
\sys{'s} \toe combined with our strategies for addressing 
the domain gap improve labeling accuracy when tested on 
held-out annotations. 
We describe our approach in Section~\ref{sec:approach} and results in Section~\ref{sec:results}.

\subsubsection{Contributions.}
We show that \sys can achieve excellent accuracy on held-out annotations. 
Our domain gap mitigation strategies allow \sys to reduce labeling noise. 
On color annotations, we can achieve over 95\% accuracy in labeling. 
We further investigate noisy self-training to iteratively improve 
labeling accuracy to 98\%. 
We show the effectiveness of the generated KID on several vehicle 
recognition problems, where we achieve state-of-the-art results 
in VMMR and vehicle re-id. 
The contributions are:

\squishitemize
\item \textbf{\sys}: We demonstrate our \toe approach that transfer 
knowledge between diverse vehicle datasets to generate a KID for multiple 
vehicle recognition tasks, such as VMMR, re-id, parts detection, and tracking.

\item \textbf{KID}: We release our knowledge-integrated dataset for vehicle 
recognition tasks, available by request\footnote{https://forms.gle/c5q5kbo62zCou7da7}. 
We hope this integrated dataset can further improve feature extraction for vehicle
recognition problems such as VMMR, re-id, tracking, etc. 
We will continue to improve this KID with any new released annotations.
\squishend

\section{Related Work}
\label{sec:related}


\subsection{Vehicle Feature Extraction}
Current work in vehicle feature extraction has 
focused on representation learning for re-id and
on fine-grained classification for vmmr.
Specialized appraches use multi-branch networks, feature fusion, attention 
modules, keypoints, guided attention, as well as synthetic images from GANS.
%
%
%
%
%
%

\subsubsection{Datasets.}
Liu et al. released VeRi-776~\cite{veri}, with 776 unique vehicle identities, 
each with images from multiple cameras in different ambient conditions. 
%
%
%
Performance is evaluated with the mAP and rank-1 metric. 
%
%
%
The BoxCars116K dataset in~\cite{boxcars} provides vehicles annotations with 
their body type.
Cars196~\cite{cars196} contains make and model attributes of vehicle images.
VColors~\cite{vcolors} contains color labels for vehicles. 
CompCars~\cite{compcars} is a fine-grained dataset with make, model, and type annotations.

\subsection{Cross-Dataset Knowledge Transfer}
Knowledge transfer has been implemented for several distinct tasks, 
such as object detection, person re-id, and image classification. 
Multiple datasets are integrated to either improve 
performance on a single task or improve cross-dataset domain transfer. 
%

%
Human labelers manually complete missing annotations~\cite{human}. 
This naturally comes at significant expense in terms of money and time. 
Active learning and human-in-the-loop methods mitigate these costs; 
randomized annotator selection and agreement is often employed to ensure 
labeling accuracy~\cite{collab}. 
%
%
Extending existing datasets to include new annotations requires 
processing each sample again. 

There are several approaches for supervised, semi-supervised, 
and unsupervised domain adaptation in object recognition~\cite{cross5,cross4}, 
as well as re-id~\cite{cross1,cross2,cross3}. 
In each case, the solution focuses on generalization between 
datasets by reducing dataset overfitting, instead of knowledge 
transfer. 
%
 Person re-id approaches use existing re-id datasets to test 
domain transfer~\cite{cross3,cross7}. 
Domain adaptation is common for object detection as well 
~\cite{unified}. 


\subsubsection{Knowledge-Integrated Datasets with ATEAM.} 
Compared to above approaches, \sys integrates distinct label 
spaces across several datasets, as shown in Figure~\ref{fig:kid}. 
Since it is purely algorithmic, \sys scales significantly 
better than manual labeling. 
The tradeoff is minimal noise in labeling. 
Recent research in~\cite{noisy} indicate slight noise is tolerable for 
most tasks, and in fact improves robustness to overfitting. 
%

\section{ATEAM}
\label{sec:approach}

We now describe \sys, our annotation \toe approach for 
knowledge-integrated datasets. 
We have shown \sys{'s} dataflow in Figure~\ref{fig:integratedoverview}.
Here, we describe the components in detail. 
First, we will set up preliminaries such as notation, 
backbone, and evaluation baselines. 
Then we cover team construction for each annotation. 
Finally, we cover domain gap mitigation strategies.

\subsection{Preliminaries}

\subsubsection{Notation.}
We work with $D$ datasets $d_1,d_2,\cdots,d_D$. 
These datasets contain vehicle images as samples, 
each labeled with some annotations. 
The datasets comprise $k$ annotations $y_1,y_2,\cdots,y_k$. 
Each dataset contains some subset of annotations. 
$\{d\}^k$ is the set of datasets that contain annotation 
$y_k$, with size $S_k$. 
$\{d^*\}^k$ is the complementary set of datasets without 
annotation $y_k$, with size $D-S_k$. 

\subsubsection{Backbone.} 
We use ResNet with BNN~\cite{bnn} as our backbone for each team member. 
This includes a bag of tricks for performance improvements, 
such as warmup learning rate and a batch-norm neck. 
We conducted preliminary experiments with ResNet-18, 34, 50, 
and 101; as expected, increased layers contribute to higher accuracy. 
Since this increase is independent relative to improvement due to 
domain gap mitigation strategies, we use ResNet-18 in our experiments 
to reduce memory costs. 
As the creation of KID is a preprocessing step, increased space and 
time complexity due to larger team members is amortized over subsequent 
task-specific model training. 
We leave extensions to larger models for future work.

\subsubsection{Evaluation Baseline.} 
We conduct 2 sets of evaluations: (i) we verify \sys{'s} cross-dataset 
labeling accuracy; and (ii) we verify the effectiveness of the 
generated Knowledge-Integrated Dataset (KID) for vehicle recognition problems. 
For \sys{'s} labeling accuracy, we evaluate accuracy for each annotation 
$k$ using a held-out dataset from $\{d\}^k$ itself; for the experiments, 
we build a team using all but one dataset from $\{d\}^k$. 
Then we compare annotation labeling accuracy to the ground truth 
in the held-out dataset. 
We perform this held-out accuracy experiment using each $d\in\{d\}^k$ 
as the held-out dataset and average the results. 

To evaluate the effectiveness of KID, we perform several vehicle 
recognition tasks using off-the-shelf, existing states-of-art 
in VMMR and re-id. 
We will show in Section~\ref{sec:results2} that the KID is an excellent 
resource for any vehicle recognition task: the knowledge integrat-ed 
from several datasets can improve accuracy on any number of related 
tasks using only off-the-shelf models. 
We also show that combining several proposals from off-the-shelf 
approaches yields new states-of-the-art for VMMR and vehicle re-id 
when trained on KID.

\subsection{Team Construction}
\label{sec:team}
First, \sys constructs a labeling team of experts to fill in missing 
annotations. 
Given the $k$-th annotations, we have $\{d\}^k$ as the subset of 
federated datasets containing the annotation, 
and $\{d^*\}^k$ as the complementary subset without the annotation. 
For each annotation, we build a team $T_k$, with members $t_j$ trained on 
$\{d\}^k$; where $|T_k|\propto|\{d\}^k|=S_k$. 
Then, $T_k$ labels the `missing' $k$-th annotation for $\{d^*\}^k$.

Let $k$ be the color annotation $y_{color}$. 
Then $T_{color}$ has three members $t_{1,2,3}$ ($|T_{color}|=S_{color}=3$), 
trained on VeRi, VColors, and CVehicles, respectively. 
During labeling, we take the $\{d^*\}^{color}$ datasets without $y_{color}$, 
and generate their color labels using weighted voting from $T_{color}$; 
$|\{d^*\}^{color}|=D-S_{color}$. 
Since each member is trained on a different dataset, $T_{color}$ is a \toe. 
So, members contribute vote with dynamic confidence weights based on 
distance, as we will describe next. 
We have shown this labeling pipeline for $y_{color}$ in Figure~\ref{fig:integratedoverview}.

During member training, gradients from each dataset in $\{d\}^k$ contributes 
to each member. 
Specifically, each $t_j\in T_k$ is initially trained on a single dataset 
$d_j\in \{d\}^k$. 
We use early stopping tuned to cross-validation accuracy across all datasets 
in $\{d\}^k$ (see Section~\ref{sec:dgm}). 
After training, we tune each $t_j$ with gradients from $\{d\}^k\\d_j$. 
This allows members to integrated knowledge within the existing annotated 
space and improve knowledge transfer between themselves, as we show in Section~\ref{sec:results}.

\subsection{Domain Gap Mitigation}
\label{sec:dgm}
While each team can transfer knowledge from $\{d\}^k$ to $\{d^*\}^k$, 
the labeling is noisy due to domain gaps. 
Domain gaps existing in cross-dataset evaluation due to differences in 
dataset distributions~\cite{odin,cross3}. 
Since source and target datasets are drawn independently, there are samples 
in the target datasets in $\{d^*\}^k$ that team members have not 
generalized to. 
For \sys, we augment domain adaptation with bootstrapping within 
members, plus confidence weights and dynamic agreement threshold mechanisms 
to  mitigate the domain gap. 
Through this, \sys further improves accuracy, as shown in Section~\ref{sec:results}, 
and detects potentially noisy misclassifications through the agreement 
threshold mechanism. 
We distill the mitigation strategies in \sys into three steps: 
(i) bootstrapping, (ii) confidence weights, and (iii) agreement threshold.

\subsubsection{Bootstrapping.}
Our goal with bootstrapping is to increase knowledge coverage of a single 
dataset and reduce overfitting. 
Bootstrapping with bagging is applied to each member in a team. 
This reduces bias and variance of the classifiers, and reduces overfitting. 
We conducted preliminary experiments over different bagging ensemble sizes and used 5 
subsets of the training data as a good tradeoff between accuracy and training time. 
During training of a member in $T_k$, we compute validation accuracy over all $\{d\}^k$. 
This allows us to evaluate generalization across datasets for the bagging ensemble. 
During training, we employ early stopping tuned to validation accuracy across 
all datasets in $\{d\}^k$~\cite{meta}.

We combine bootstrap aggregation with multiple compression levels for each model 
in the bagging ensemble. 
Specifically, each model in the bagging ensemble for a team member 
provides predictions over different compression levels. 
Compression dampens the high-frequency region of an image; 
accordingly, overfitting and the subsequent domain gap from dataset-specific 
artifacts are mitigated. 
Using the approach from~\cite{shield}, each model in the bagging ensemble takes 4 
copies of an unlabeled image: 1 image with no compression, and 3 
images compressed with the JPEG protocol, with quality factors 90, 70, and 50, 
respectively. 
The model provides the majority predicted label from the 4 compression levels. 
The JPEG compressions are used only during prediction. 

\subsubsection{Member Confidence.}
During prediction, \sys dynamically adjusts the weight of each team member 
based on confidence. 
We compute confidence as inverse weighted distance between sample and 
training data. 
Given a sample $x_i$ to be labeled with annotation $k$, first, \sys 
computes a label $y_{ij}$ from each team member $t_j\in T_k$ 
(i.e. there are $j$ predictions for $y_i|x_i$, one for each team member 
in $T_j$). 
Then, each of the $j$ predictions for $y_i$ is weighted by the distance 
between $x_i$ and the corresponding training data center $c_j$ for each $t_j$. 
This allows members with closer training data to have higher impact 
on the predicted label for $x_i$. 
We compute the weight for each member $t_j$ given a sample $x_i$ as

\begin{align}
w_j & = \frac{1}{S_k} \big( 1 - \frac{l_2 (x_i, c_j)}{\sum_j^{S_k} l_2 (x_i, c_j)}\big)
\end{align}

where $l_2()$ is the Euclidean distance.

\subsubsection{Agreement Threshold.}
However, weighting only on distance to training data can lead to 
accuracy degradation if nearby models are over-confident in incorrect predictions. 
So, we adjust the majority voting threshold instead of maintaining it 
at $>0.5$. 
Specifically, we increase this agreement threshold by $\alpha$ for each $x_i$, 
where $\alpha$ corresponds to the divergence between the 
dataset-confidence weights and some predetermined initial weights 
of each team member. 
We compute the initial weights $q_j$ for each member $t_j$ using the 
cross-dataset validation accuracy from training by using the weighted probabilistic 
ensemble from~\cite{cawpe}. 
Given cross-validation f-score $f_j$ and $\beta=4$ per the CAWPE algorithm in~\cite{cawpe}, 
we compute $q_j$ as

\begin{align}
    q_j = \frac{f_j^{\beta}}{\sum_j^{S_k} f_j^{\beta}} & .
\end{align}

Then, for each $x_i$, we compute the divergence as the relative entropy, 
or KL-divergence, between $q_j$ and the dynamic confidence weight distribution. 
Then, we can use this to adjust the threshold adjustment $\alpha_k$  for team $T_k$ for a sample $x_i$, with

\begin{align}
    \alpha_k = 0.5 (1 - \exp (D_{KL} (P||Q) )) = 0.5\big( 1 - \exp ( \sum_j^{S_k} p_j \log \frac{p_j}{q_j}) \big) &.
\end{align}

Since KL-divergence increases without bound, we map it to $[0,1]$. 
Here, $p_j=w_j$ from Eq 1, which is the normalized dataset-distance weight 
between a sample $x_i$ and member $t_j$'s training data center $c_j$.

\begin{figure}[t]
    \centering
    \includegraphics[width=0.7\textwidth]{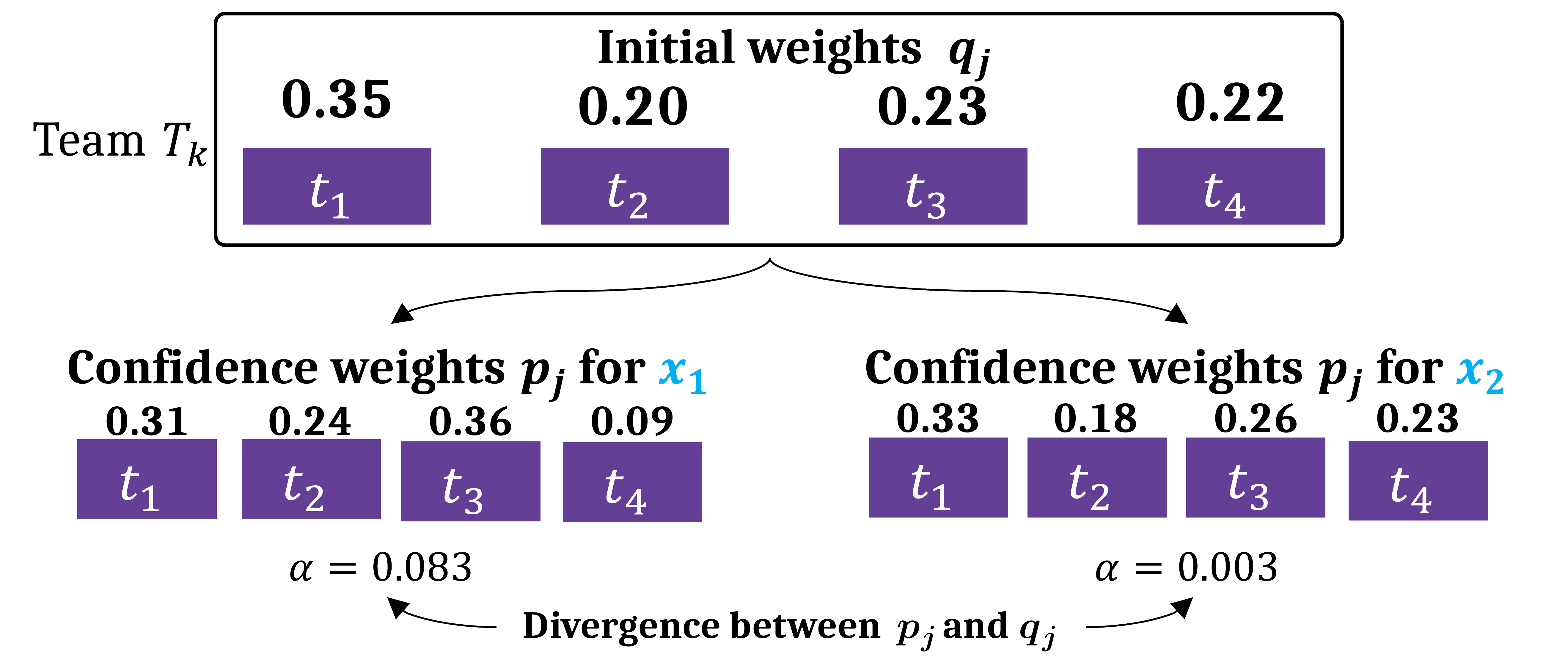}
    \caption{An example of agreement threshold adjustment using KL-divergence
    between initial weights and dynamic confidence weights. Initial weights are
    obtained with normalized cross-dataset accuracy.}
    \label{fig:agreement}
\end{figure}

Our classification threshold for label prediction is therefore $0.5+\alpha$. 
If a sample's labeling agreement does not achieve the threshold, we ignore predictions 
and leave the $k$-annotation blank for that sample. 
We show an example of threshold adjustment in Figure~\ref{fig:agreement} for a single team. 
Here, the first sample $x_1$ is close to $t_1$ and $t_3$, and very far from $t_4$, 
yielding a confidence weight distribution with a relatively high divergence 
from the baseline weights. 
This yields a classification threshold adjustment of $\alpha=0.083$ for $x_1$, 
with a classification threshold of $0.583$. 
By comparison, $x_2$ has a relatively closer weight distribution, so $\alpha=0.003$, 
with threshold $0.503$. 
The classification thresholds, applied to each $x_i$, allows annotation teams to 
achieve higher accuracies; the color team achieves an accuracy of 0.98, 
as we show in Table~\ref{tab:color}.

\setlength{\tabcolsep}{4pt}
\begin{table}[t]
    \centering
    \caption{Datasets: We integrated the following datasets to create the KID. Vehicle-KID contains all images from the datasets, as well as annotations
    labeled with the respective teams.}
    \label{tab:datasets}
    \begin{tabular}{lrrrrr}
    \hline
Dataset                         & Color                     & Type                          & Make                        & Model                 & Samples     \\ \hline
    CompCars~\cite{compcars}    & No                        & Yes            & Yes                & Yes          & 136K        \\
    BoxCars116K~\cite{boxcars}  & No                        & No          & Yes       & Yes & 116K        \\
    Cars196~\cite{cars196}      & No                        & No          & Yes       & Yes & 16K         \\
    VColors~\cite{vcolors}      & Yes                       & No                      & No                    & No                    & 10K         \\
    VeRi~\cite{veri}            & Yes                       & Yes                     & No                    & No              & 50K         \\
    CVehicles                   & Yes                       & Yes                     & No                    & No              & 165K        \\ \hline
    Vehicle-KID                 & Yes                       & Yes                   & Yes                    & Yes              & $\sim$500K  \\ \hline
    \end{tabular}
    \end{table}

\subsection{Experimental Setup}

\subsubsection{System Details.}
We implemented \sys in PyTorch 1.10 on Python 3.10. 
Each member in \sys uses a ResNet-18 backbone. 
Our experiments are performed on a server with NVIDIA Tesla K40 GPU, 
and an Intel Xeon 2GHz processor, with datasets stored on a remote drive.

\subsubsection{Datasets.} 
We use the following datasets: CompCars~\cite{compcars}, Box-Cars116K~\cite{boxcars}, Cars196~\cite{cars196}, 
VColors~\cite{vcolors}, and VeRi~\cite{veri}. 
We also obtained our own dataset of vehicles labeled with color and 
type annotations using a web crawler on a variety of car-sale sites, 
called CVehicles. 
Datasets are described in Table~\ref{tab:datasets}. 
We use CompCars, BoxCars116K, and Cars196 for end-to-end evaluations of VMMR, and VeRi for re-id evaluation. 
Their annotations are incomplete, since none contain all three desired annotations of color, type, and make. 

\section{ATEAM Evaluation}
\label{sec:results}
We first show the effectiveness \sys{'s} annotation and 
domain mitigation strategies. 
To evaluate, we build three 
annotation teams: Color-TEAM, Type-TEAM, and Make-TEAM. 
The Color Team is built on VColors, VeRi, and CVehicles. 
The Type Team is built with CompCars, CrawledVehicles, and VeRi. 
Finally, the Make Team is built with CompCars and CVehicles. 
Additionally, since `makes' are more diverse than color or type, we ensured that CVehicles
contiains a superset of the CompCars make annotations.
To examine the effect domain gap mitigation strategies, we use each 
dataset's test set in the cross-dataset acccuracy calculation and average 
the results for each team. 
During training of each team member, we employ common data augmentations such 
as flipping, cropping, and random erasing. 
We also employ the warmup learning rate from~\cite{bnn}.

For each team, we evaluate with a baseline first that uses no 
domain gap mitigation strategies. The baseline is constructed using a ResNet backbone.
Then, we add and successively test each of the following strategies: bootstrapping, 
early stopping, JPEG compression, confidence weights, and agreement thresholds. 
For JPEG Compression, we compare between a single additional compression 
of qf=90, and set of 3 compressions at qf=$\{90,70,50\}$.

\begin{table}[t]
    \centering
    \caption{Color Team: Using our domain gap mitigation strategies, we can improve cross-dataset 
    color labeling accuracy by 12\%. Each of our mitigation strategies contributes to improved 
    out-of-distribution generalization.}
    \label{tab:color}
    \begin{tabular}{lrrr}
    \hline
    \textbf{Strategy}              & \textbf{VColors} & \textbf{VeRi} & \textbf{CVehicles} \\ \hline
    Initial                        & 0.87             & 0.84          & 0.86               \\
    +Bootstrap                     & 0.88             & 0.87          & 0.89               \\
    +Early   Stop                  & 0.89             & 0.90          & 0.92               \\
    +Compression (90)              & 0.91             & 0.91          & 0.92               \\
    +Compression(90, 70, 50)       & 0.94             & 0.93          & 0.94               \\
    +Confidence   Weights          & 0.95             & 0.95          & 0.96               \\
    \textbf{+Agreement  Threshold} & \textbf{0.98}    & \textbf{0.97} & \textbf{0.98}      \\ \hline
    \end{tabular}
    \end{table}

\subsubsection{Color Team.}
We show results in Table~\ref{tab:color}.
For the Color Team, the initial cross-dataset accuracy is $\sim 0.86$. 
This is due to the domain gaps: while color detection is relatively straightforward, 
domain gaps reduce accuracy on the cross-dataset evaluation. 
With bootstrapping and early stopping, we can improve accuracy to 0.89. 
Adding JPEG compression, with 3 compression levels, 
further increases accuracy another 5\%, to 0.95. 
Compression mitigates high-frequency differences between the dataset images. 
In conjunction with confidence weights and KL-divergence based agreement threshold 
adjustment, we can push color labeling accuracy to 0.98. 
In generating the KID, we employ this team to label the remaining datasets 
that do not have color annotations (Cars196, BoxCars, CompCars). 
For each sample, when the majority label's weights do not meet the sample's 
agreement threshold, that color annotation is left blank. 
During training, we address these blank annotations with the 
subset training method from~\cite{mtml}.

\begin{table}[t]
    \centering
    \caption{Type Team: Similar to Color Team, mitigation strategies are instrumental in improve cross-dataset generalization for type labeling.}
    \label{tab:type}
    \begin{tabular}{lrrr}
    \hline
    \textbf{Strategy}               & \textbf{CompCars} & \textbf{CVehicles} & \textbf{VeRi} \\ \hline
    Initial                         & 0.86              & 0.88               & 0.86          \\
    +Bootstrap                      & 0.87              & 0.89               & 0.88          \\
    +Early   Stop                   & 0.89              & 0.91               & 0.89          \\
    +Compression   (90)             & 0.91              & 0.91               & 0.91          \\
    +Compression(90,   70, 50)      & 0.93              & 0.94               & 0.94          \\
    +Confidence Weights             & 0.96              & 0.95               & 0.95          \\
    \textbf{+Agreement   Threshold} & \textbf{0.98}     & \textbf{0.97}      & \textbf{0.98} \\ \hline
    \end{tabular}
    \end{table}

\subsubsection{Type Team.}
We show results in Table~\ref{tab:type}.
Similarly, for the Type Team, we have an initial accuracy of $\sim 0.87$. 
We improve this to 0.89 with bootstrap aggregation and early stopping. 
Using JPEG compression further increases accuracy to 0.93. 
Finally, confidence weights and agreement thresholds lead to accuracy of 
0.98 in cross-dataset type labeling, which is a 14\% increase over the baseline team's 
cross-dataset accuracy.

\begin{table}[t]
    \centering
    \caption{Make Team: For make detection, we use triplet mining and clustering due to annotation diversity. Accuracy is computed as fraction of makes correctly detected. 
    }
    \label{tab:make}
    \begin{tabular}{lrr}
    \hline
    \textbf{Strategy}               & \textbf{CompCars} & \textbf{CVehicles} \\ \hline
    Initial                         & 0.84              & 0.83               \\
    +Bootstrap                      & 0.86              & 0.87               \\
    +Early   Stop                   & 0.87              & 0.87               \\
    +Compression   (90)             & 0.88              & 0.88               \\
    +Compression(90,   70, 50)      & 0.89              & 0.89               \\
    +Confidence Weights             & 0.90              & 0.91               \\
    \textbf{+Agreement   Threshold} & \textbf{0.92}     & \textbf{0.92}      \\ \hline
    \end{tabular}
    \end{table}

\subsubsection{Make Team.}
For make classification with the Make Team, we train with the triplet loss setting.
This is because vehicle makes are more diverse than color and type, and each dataset has
only subsets of all make labels.
In this case, learning to cluster is more effective than labeling with pre-determined
classes that may not appear in the cross-dataset setting.
We have an initial accuracy of ~0.84 averaged across the CompCars and CVehicles datasets; this is computed by the
fraction of makes we are able to cluster. 
We also first ensured that CVehicles contains the makes from CompCars. 
With all strategies working together, we can achieve a ~9\% increase in 
accuracy in the cross-dataset labeling, from $0.84$ to $0.92$.

\subsubsection{KID Generation}
We then use these teams to complete annotations in the unlabeled datasets. 
Using dynamic agreement threshold can leave samples unlabeled. 
In the KID, we therefore have a fraction of each component dataset that remains unlabeled. 
During evaluation of vehicle recognition tasks trained on KID, we use the annotation subset 
training method from~\cite{mtml} for these missing annotations. 
For make annotation, we employ the triplet mining method common in re-id to instead 
label the vehicles with make clusters. 
This is because while color and type are consistent across datasets, 
make annotations are diverse. 
So, we label them as clusters for each dataset. In the next section, we discuss 
using KID for common vehicle recognition tasks such as VMMR or re-id.

\section{KID Evaluation}
\label{sec:results2}

\begin{figure}
    \centering
    \begin{subfigure}[b]{0.49\textwidth}
        \centering
        \includegraphics[width=\textwidth]{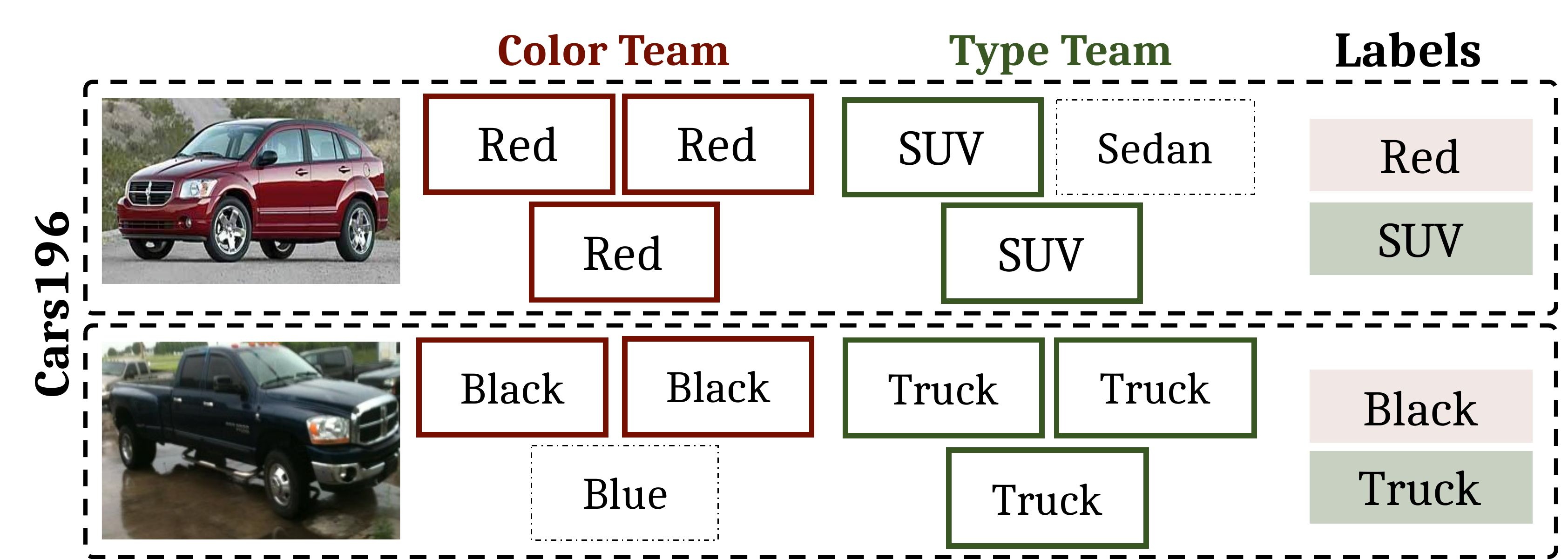}
        \caption{Two images from Cars196 labeled with color and type annotations in KID.}
        \label{fig:cars196}
    \end{subfigure}
    \hfill
    \begin{subfigure}[b]{0.49\textwidth}
        \centering
        \includegraphics[width=\textwidth]{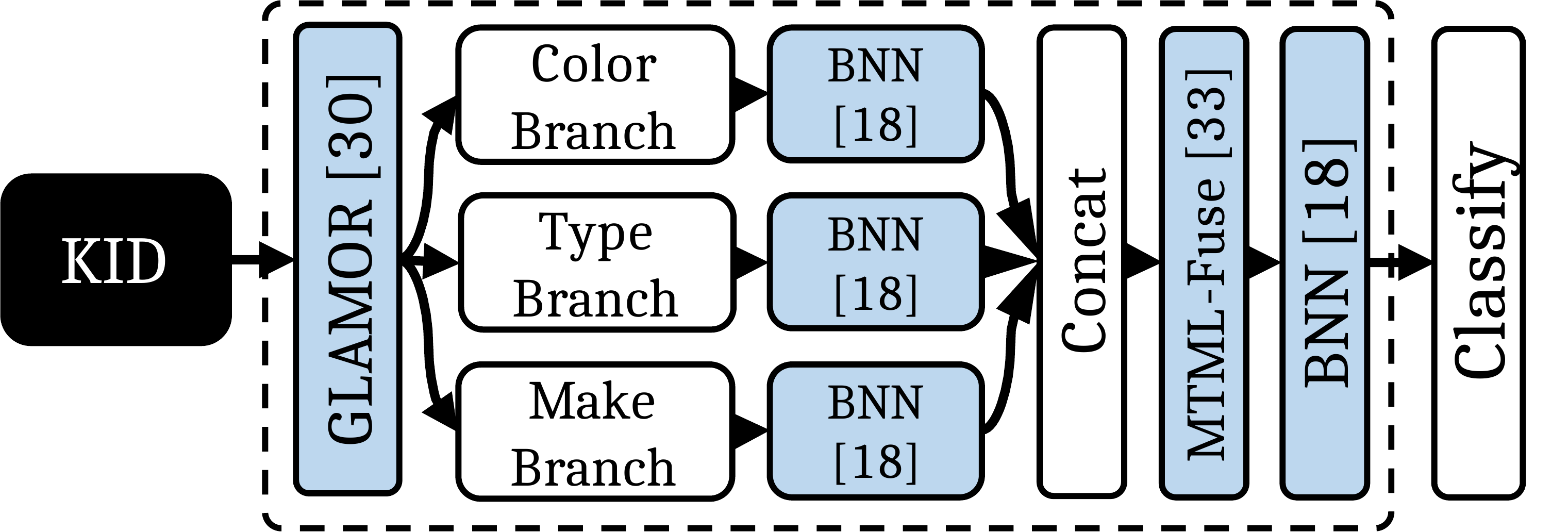}
        \caption{Prototype model for evaluating KID. Blue components are existing specializations for vehicle feature extraction.}
        \label{fig:model}
    \end{subfigure}
       \caption{Labels in KID, and model trained on these labels for VMMR and re-id.}
       \label{fig:kidmodel}
\end{figure}

Using the domain mitigation strategies, we have built our 
knowledge-integrated dataset (KID). 
KID includes all three annotations (color, type, and make) 
for all images in its dataset. 
%
%
As such, KID is a union over the distinct federated datasets 
that comprise our experiments. We show an example of labels in KID 
in Figure~\ref{fig:cars196}.  

\subsection{Vehicle Recognition Problems}
We now show the impact of knowledge transfer from federated datasets 
with several models trained on the KID. 
As we have mentioned, each vehicle recognition problem can take advantage 
of the data in every vehicle recognition dataset, specifically image features. 
The bottleneck is usually manually labeling missing annotations in existing datasets. 
With \sys, we create the KID that bypasses these issues.

To evaluate KID, we train models for vehicle re-id and VMMR. 
Usually, such models use different datasets: VeRi is commonly used for re-id, 
and CompCars is commonly used for VMMR. 
In KID, both datasets are integrated with all desired annotations, 
allowing us to directly use them for both VMMR and re-id.

\begin{table}[t]
    \caption{Re-ID: We train off-the-shelf re-id models on VeRi and KID, and test on VeRi. Using KID increases mAP compared to using VeRi due to increased knowledge from the federated datasets.}
    \label{tab:reid}
    \begin{tabular}{|l|rlrrlrrlr|}
    \hline
    \multicolumn{1}{|c|}{\multirow{3}{*}{Approach}} & \multicolumn{9}{c|}{Trained on VeRi vs Trained on KID}                                                                                                                                                                                                             \\ \cline{2-10} 
    \multicolumn{1}{|c|}{}                          & \multicolumn{3}{c|}{mAP}                                                         & \multicolumn{3}{c|}{Rank@1}                                                      & \multicolumn{3}{c|}{Rank@5}                                           \\ \cline{2-10} 
    \multicolumn{1}{|c|}{}                          & \multicolumn{1}{c}{VeRi} & $\rightarrow$ & \multicolumn{1}{c|}{KID}            & \multicolumn{1}{c}{VeRi} & $\rightarrow$ & \multicolumn{1}{c|}{KID}            & \multicolumn{1}{c}{VeRi} & $\rightarrow$ & \multicolumn{1}{c|}{KID} \\ \hline
    OIFE+ST~\cite{oife}                                         & 51.42                    & $\rightarrow$ & \multicolumn{1}{r|}{-}              & 68.30                    & $\rightarrow$ & \multicolumn{1}{r|}{-}              & 89.70                    & $\rightarrow$ & -                        \\
    Hard-View-EALN~\cite{ealn}                                 & 57.44                    & $\rightarrow$ & \multicolumn{1}{r|}{-}              & 84.39                    & $\rightarrow$ & \multicolumn{1}{r|}{-}              & 94.05                    & $\rightarrow$ & -                        \\ \hline
    Baseline                                   & 72.15                    & $\rightarrow$ & \multicolumn{1}{r|}{75.21}          & 91.65                    & $\rightarrow$ & \multicolumn{1}{r|}{93.61}          & 96.91                    & $\rightarrow$ & 98.34                    \\
    BNN~\cite{bnn}                                       & 77.15                    & $\rightarrow$ & \multicolumn{1}{r|}{80.14}          & 95.65                    & $\rightarrow$ & \multicolumn{1}{r|}{96.13}          & 97.91                    & $\rightarrow$ & 98.59                    \\
    MTML~\cite{mtml}                                            & 68.30                    & $\rightarrow$ & \multicolumn{1}{r|}{73.61}          & 92.00                    & $\rightarrow$ & \multicolumn{1}{r|}{{\ul 96.87}}    & 94.20                    & $\rightarrow$ & 98.23                    \\
    GLAMOR~\cite{glamor}                                          & 80.34                    & $\rightarrow$ & \multicolumn{1}{r|}{81.65}          & 96.53                    & $\rightarrow$ & \multicolumn{1}{r|}{96.78}          & 98.62                    & $\rightarrow$ & {\ul 98.67}              \\ \hline
    \textbf{BMG-ReID}                               & {\ul 81.86}              & $\rightarrow$ & \multicolumn{1}{r|}{\textbf{83.41}} & 96.47                    & $\rightarrow$ & \multicolumn{1}{r|}{\textbf{97.12}} & 98.47                    & $\rightarrow$ & \textbf{98.94}           \\ \hline
    \end{tabular}
    \end{table}

\subsubsection{Re-ID.}
With the KID as the training data, we use the architectures presented in~\cite{glamor,mtml,bnn}. 
Since the integrated dataset has annotations for color, make, and type, 
we leverage these annotations in a multi-branch model, similar to~\cite{mtml,ref}, 
while using the VeRi subset to guide re-id feature learning. 
We show a prototypical model in Figure~\ref{fig:model}.
Each branch performs annotation-specific feature extraction. 
As in~\cite{mtml}, annotation labels act as branch-specific targets. 
The subset-training method from~\cite{mtml} allows us to `skip' samples that do not have 
an annotation label. 
We then fuse branch features for re-id. We describe the variations below, and results in 
Table~\ref{tab:reid}:

\squishitemize
\item \textbf{Baseline}: The baseline re-id model uses branches for color, 
type, and make. 
The blue components in Figure~\ref{fig:model} are not used. 
Features are concatenated for re-id. 
This model achieves similar accuracy to several existing benchmarks, 
with mAP of 0.75 when trained on KID, and rank-1 of 93\%, indicating the knowledge transfer 
into the integrated dataset is competitive with specialized re-id architectures.

\item \textbf{BNN}: After branch-specific features, a BNNeck~\cite{bnn} 
is used to improve feature normalization. This normalization allows features to cluster more discretely. BNNeck improves mAP to 0.80

\item \textbf{GLAMOR}: The backbone employs attention modules from GLAMOR~\cite{glamor}
to improve feature extraction. We use both global and local attention from~\cite{glamor}. 
We can improve performance here as well, with mAP of 0.82 on VeRi.

\item \textbf{MTML}: For feature fusion, we employ the consensus loss in~\cite{mtml} 
to propagate soft-targets to each branch. This also improves mAP from the baseline to 0.81

\item \textbf{BMG-ReID}: We combine the above approaches' key contributions. 
BMG-ReID uses a BNNeck~\cite{bnn}, attention modules~\cite{glamor}, and consensus loss for fusion~\cite{mtml}. 
To better evaluate BMG-ReID, we compare to BMG, which is trained on VeRi only. 
As we see, BMG itself improves on the baseline, from 0.79 to 0.81. With BMG-ReID, 
we further improve to state-of-the-art results due to the additional knowledge, 
with mAP of 0.83 and Rank-1 of 0.96.

\squishend

\subsubsection{VMMR.}
We evaluate the same models for VMMR: a multiple branches baseline and variations 
that integrate key contributions from related work.
The models are evaluated on CompCars. 
We present the variations below and show results in Table~\ref{tab:vmmr}:

\squishitemize
\item \textbf{Baseline}: Similar to the re-id baseline, we use three branches. 
Branch features are concatenated and used for fine-grained VMMR classification. 
This achieves accuracy of 0.91

\item \textbf{BNN}: Using BNNeck~\cite{bnn} improves accuracy to 0.92 
with improved feature normalization and projection.

\item \textbf{GLAMOR}: Global and local attention from~\cite{glamor} improve accuracy to 0.92 from the baseline. 

\item \textbf{MTML}: With~\cite{mtml}'s consensus loss we can achieve accuracy of 0.94. 
Similar to the re-id counterpart, the consensus loss ensures features are projected to similar semantic space.

\item \textbf{BMG-VMMR}: We then combine the key contributions of the above models 
to achieve accuracy of 0.96 on CompCars. 
With a backbone of ResNet50, we can increase this to 0.97. 
This is comparable to D161-SMP~\cite{vmmr3}. Compared to D161-SMP, we use a third fewer layers and can train faster. 
We expect replacing the backbone with ResNet152 or DenseNet would further improve accuracy 
and leave exploration of other architectures to future work.

\squishend

\begin{table}[t]
    \caption{VMMR: Similar to re-id, we use off-the-shelf models, and evaluate by training on the VMMR dataset or KID, and testing on the VMMR dataset. Using KID increases accuracy compared to using a single VMMR dataset.}
    \label{tab:vmmr}
    \begin{tabular}{|l|rcrrcrrcr|}
    \hline
    \multicolumn{1}{|c|}{\multirow{2}{*}{Approach}} & \multicolumn{9}{c|}{Trained on single dataset vs Trained on KID}                                                                                                                                                      \\ \cline{2-10} 
    \multicolumn{1}{|c|}{}                          & CompCars   & $\rightarrow$ & \multicolumn{1}{r|}{KID}           & BoxCars    & $\rightarrow$ & \multicolumn{1}{r|}{KID}           & Cars196    & $\rightarrow$ & KID           \\ \hline
    R50~\cite{bnn}                                         & 0.90       & $\rightarrow$ & \multicolumn{1}{r|}{-}             & 0.75       & $\rightarrow$ & \multicolumn{1}{r|}{-}             & 0.89       & $\rightarrow$ & -             \\
    D161-SMP~\cite{vmmr3}                                        & {\ul 0.97} & $\rightarrow$ & \multicolumn{1}{r|}{-}             & -          & $\rightarrow$ & \multicolumn{1}{r|}{-}             & {\ul 0.92} & $\rightarrow$ & -             \\ \hline
    Baseline                                   & 0.83       & $\rightarrow$ & \multicolumn{1}{r|}{0.91}          & 0.71       & $\rightarrow$ & \multicolumn{1}{r|}{0.88}          & 0.82       & $\rightarrow$ & 0.89          \\
    BNN~\cite{bnn}                                       & 0.88       & $\rightarrow$ & \multicolumn{1}{r|}{0.92}          & 0.79       & $\rightarrow$ & \multicolumn{1}{r|}{0.89}          & 0.86       & $\rightarrow$ & 0.91          \\
    MTML~\cite{mtml}                                            & 0.88       & $\rightarrow$ & \multicolumn{1}{r|}{0.92}          & 0.82       & $\rightarrow$ & \multicolumn{1}{r|}{0.88}          & 0.85       & $\rightarrow$ & 0.90          \\
    GLAMOR~\cite{glamor}                                          & 0.91       & $\rightarrow$ & \multicolumn{1}{r|}{0.94}          & 0.85       & $\rightarrow$ & \multicolumn{1}{r|}{0.89}          & 0.88       & $\rightarrow$ & 0.91          \\ \hline
    \textbf{BMG-VMMR}                               & 0.96       & $\rightarrow$ & \multicolumn{1}{r|}{\textbf{0.97}} & {\ul 0.91} & $\rightarrow$ & \multicolumn{1}{r|}{\textbf{0.92}} & 0.91       & $\rightarrow$ & \textbf{0.93} \\ \hline
    \end{tabular}
    \end{table}

\section{Conclusion}
In this paper, we have presented \sys, a \toe approach for 
knowledge transfer between federated datasets of related deep 
learning problems. 
\sys generates an integrated dataset that is useful for the union 
of all deep learning problems represented by federated datasets. 
We evaluate \sys for vehicle recognition problems and show that our integrated 
dataset can help off-the-shelf models achieve excellent accuracy on 
VMMR and vehicle re-id with no changes to model architectures. 
We also show that adjusting models to take advantage of \sys{'s} diverse 
annotations further improves accuracy, with new states-of-the-art on 
VMMR and re-id: we achieve accuracy of 0.97 for VMMR on CompCars, and mAP 0.83
for re-id on VeRi. 
We release both the \sys code and our knowledge-integrated dataset KID 
for vehicle recognition for public use.

The improvements due to \sys suggest significant research opportunities 
and practical applications. 
For example, the knowledge integration approach may be applicable to 
other problem domains where schema incompatibility has encouraged the 
creation of related, but non-overlapping datasets on subdomains. 
Another example of challenging problems that would be amenable to 
knowledge integration approach consists of recognizing flexibly 
composed vehicles such as semi-trailer trucks, 
where the tractor unit may be coupled to a variety of trailer units.

\clearpage
\bibliographystyle{splncs04}
\bibliography{egbib}
\end{document}